\documentclass[letterpaper, 10pt, conference]{ieeeconf}
\pdfoutput=1
\let\chapter\section
\usepackage[linesnumbered,ruled]{algorithm2e}
\def\BState{\State\hskip-\ALG@thistlm}
\usepackage{pifont}
\usepackage{booktabs}
\newcommand{\ra}[1]{\renewcommand{\arraystretch}{#1}}
\usepackage{multirow}
\usepackage{graphicx}
\usepackage{amsfonts}
\usepackage{amsmath,amssymb,mathrsfs}
\usepackage{array}
\usepackage{flafter}
\usepackage{cite}
\usepackage[tight]{subfigure}
\usepackage{verbatim}
\usepackage{color}
\usepackage{psfrag}
\usepackage{umoline}
\usepackage{hyperref}
\usepackage{indentfirst}

\newtheorem{lemma}{Lemma}
\newtheorem{example}{Example}

\allowdisplaybreaks[1]
\makeindex             
\providecommand{\keywords}[1]{\textbf{\textit{Index terms---}} #1}
\title{Feedback Motion Planning Under Non-Gaussian Uncertainty and Non-Convex State Constraints}

\author
{
Mohammadhussein Rafieisakhaei, Amirhossein Tamjidi, Suman Chakravorty, P. R. Kumar\\
Texas A\&M University
}

\begin{document}

\maketitle
\vspace{-15pt}
\begin{abstract}
Planning under process and measurement uncertainties is a challenging
problem. In its most general form it can be modeled as a Partially
Observed Markov Decision Process (POMDP) problem. However POMDPs are
generally difficult to solve when the underlying spaces are continuous,
particularly when beliefs are non-Gaussian, and the difficulty is
further exacerbated when there are also non-convex constraints on
states. Existing algorithms to address such challenging POMDPs are
expensive in terms of computation and memory. In this paper, we
provide a feedback policy in non-Gaussian belief space via solving
a convex program for common non-linear observation models. The
solution involves a Receding Horizon Control strategy using particle
filters for the non-Gaussian belief representation. We develop a way of
capturing non-convex constraints in the state space and adapt the
optimization to incorporate such constraints, as well. A key advantage of this
method is that it does not introduce additional variables in the
optimization problem and is therefore more scalable than existing
constrained problems in belief space. We demonstrate the performance
of the method on different scenarios.
\end{abstract}

\section{Introduction}\label{sec:intro}
Planning under uncertainty is a challenging problem for systems with partial observability. Many problems in robotics are in this category where there is an inherent uncertainty in the measurements obtained from sensors in addition to the uncertainty in the robot's very motion. The main challenge is that controller's knowledge about the true state of the system is limited to the conditional probability distribution of the state, called the belief. Belief is a sufficient statistic \cite{Kumar-book-86} that encodes all the information available to the controller including the history of actions and observations. The space of all possible beliefs, called the belief space, is very high dimensional in practical problems and the problem consequently suffers from the curse of dimensionality \cite{Kaelbling98}, \cite{Zhou01}. This is one of the reasons that it is extremely difficult to solve Partially Observable Markov Decision Processes (POMDPs) \cite{Astrom65}, \cite{Smallwood73}, \cite{Kaelbling98} which provide the general framework for planing under uncertainty. They suffer from the curse of history \cite{Pineau03} due to the exponential growth of decision choices because of dependency of future decisions on previous ones. 

Major point-based POMDP solvers such as PBVI, HSVI, Perseus, and SARSOP (c.f. the survey in \cite{GShaniJPineau13}) consider finite state, observation and action spaces and develop a decision tree that can exactly solve the POMDP problem for the initial belief state. Recent point-based solvers such as MCVI \cite{HsuMCVI11,HsuMCVIMacro11} can allow continuous state space. However, generally in point-based solvers, the time complexity of the algorithms grows exponentially with the number of (sampled) states and time horizon \cite{Papadimitriou87,Madani99}. Further, they guarantee optimality of their solution only for the particular initial belief state. This means that if there is a deviation from the planned trajectory during the execution, it becomes impractical to re-plan and compensate for the accumulated errors due to the stochastic nature of the system. Therefore, these methods are not suitable for use in the control strategies such as Receding Horizon Control (RHC) \cite{Platt11,Platt10ML,Platt-Cvx-RHC-12} that require a fast re-planning algorithm.

Feedback-based Information RoadMap (FIRM) \cite{Ali11-FIRM-IROS}, \cite{Ali14} is a general framework to overcome the curse of history that attempts to solve an MDP in the sampled belief space. The graph-based solution of FIRM introduced an elegant method for solving POMDPS with continuous underlying spaces. However, attention is restricted to Gaussian \cite{NRoy08,NRoy11,Vitus11,van2012motion} belief spaces which can be insufficient in some problems.

Another issue is that in the planning stage, observations are random variables, and so therefore are future beliefs, which is a major challenge for computation of the solution. One approach samples the observations, generates the corresponding beliefs, and performs a Monte Carlo analysis \cite{beskos2009monte}. Unfortunately, this has a heavy computational burden, which limits its usage. Another approach restricts attention to the maximum likelihood observation \cite{Platt10ML}, and propagates the corresponding belief using the filtering equations (either linear Gaussian or non-Gaussian). This method is computationally better than the former approach, however, it still requires belief updates using the Bayesian filtering equations. In all these methods, based on a predicted observation, the belief state is propagated, and a new observation is predicted, and the control policy is thereby determined.

In this paper, we propose a different approach to this problem. Essentially, we propose taking samples from the initial belief and propagating them via a noiseless system model. Then, we use the observation model's properties to guide the samples towards the regions where the prediction of the observation error is reduced. What this means is that, essentially, we attempt to control the system towards a subset of the state space where we predict that the dispersion of the ensemble of the observation particles is reduced and therefore the beliefs are more informative.

Therefore, mathematically, we provide a general framework to control a system under motion and sensing uncertainties using an RHC strategy. We define a basic convex planning problem that can be used to obtain an optimized trajectory for a robot. Then, we utilize the knowledge of filtering to define a general cost function that incorporates the state-dependent linearized observation model to follow a trajectory along which the system gains more information about its state from the environmental features in a trade-off with the control effort along the path. We use particle representation of belief and our belief space can be non-Gaussian. The convex formulation of the problem enables us to avoid the Dynamic Programming \cite{Sniedovich06} over the belief space which is usually highly costly (in fact, intractable in exact sense). In addition, in case that the agent deviates from its planned path, due to the convexity of planning problem, we can stop execution and re-plan to obtain a better path towards the goal region. The execution of the actions stop whenever the probability of reaching to a goal region is higher than some predefined threshold.

Moreover, we develop a method to adapt the optimization problem such that the non-convex constraints on the feasible state space are respected softly. Particularly, we propose a special form of penalty functions to incorporate the obstacles into the optimization cost, which enables us to respect the constraints without addition of any new variable to the optimization problem. Therefore, the optimization remains scalable to higher number of samples and longer time horizons as opposed to the sampling based chance-constraint methods such as \cite{van1969shaped,blackmore2010probabilistic}. Finally, we run the algorithm and show the simulation results to support our work.

\section{System Representation}\label{sec:Problem}
In this subsection, we specify the required ingredients that we use for the problem.

\emph{State space definition:} For simplicity we denote the random variables with upper case and the values with lower case letters. Throughout the paper, vectors $\mathbf{x}\in \mathbb{X} \subset\mathbb{R}^{n_x}$, $\mathbf{u}\in \mathbb{U}\subset\mathbb{R}^{n_u}$, and $\mathbf{z}\in \mathbb{Z}\subset\mathbb{R}^{n_z}$ denote the state of the system, the action, and observation vectors, respectively.

\emph{System equations:} The dynamics of the system and observations are as follows:
\begin{subequations}\label{eq:linear-sys-general-app}
\begin{alignat}{2}
\!\!\!\mathbf{x}_{t+1}&\!\!=f(\mathbf{x}_{t},\mathbf{u}_{t},\boldsymbol{\omega}_{t})\label{eq:linear-sys-general-app-1}\\
\!\!\!\mathbf{z}_{t}&\!\!=h(\mathbf{x}_{t})+\boldsymbol{\nu}_{t}.\label{eq:linear-sys-general-app-2}
\end{alignat}
\end{subequations}
where $ \{\boldsymbol{\omega}_t\} $ and $ \{\boldsymbol{\nu}_t\} $ are two zero mean independent, identically distributed (iid) random sequences, and are mutually independent. In addition, $f:\mathbb{X}\times\mathbb{U}\times\mathbb{R}^{n_x}\rightarrow\mathbb{X}$ shows the process dynamics (motion model), and $h:\mathbb{X}\rightarrow\mathbb{Z}$ denotes the observation model.

\textit{Belief:} Belief or information state at time step $ t $ is a function defined as $ b_t:\mathbb{X}\times\mathbb{Z}^{t}\times\mathbb{U}^{t}\rightarrow\mathbb{R} $, where $ b_{t}(\mathbf{x},\mathbf{z}_{0:t},\mathbf{u}_{0:t-1}, b_0):=p_{\mathbf{X}_t|\mathbf{Z}_{0:t};\mathbf{U}_{0:t-1}}(\mathbf{x}|\mathbf{z}_{0:t};\mathbf{u}_{0:t-1};b_0) $, which is the posterior distribution of $ \mathbf{x}_t $ given the data history up to time $ t $ and the initial belief, $ b_0 $. In addition, the space of all possible belief states or the belief space is denoted by $ \mathbb{B} $. For simplicity of notation, we will denote the belief state with $ b_t(\mathbf{x}) $ or $ b_t $ throughout this paper \cite{Sondik71,Kumar-book-86,Bertsekas07,Thrun2005}.

\emph{Particle representation of belief:} We consider a non-Gaussian model of the belief space and approximate the belief state $b_{t}$ at time step $ t $ by $ N $ number of particles in the state space $\{\mathbf{x}^i_t\}_{i=1}^{N}$ with importance weights $\{w^i_t\}_{i=1}^{N}$ \cite{Thrun2005,CrissanDoucet02,Doucet01Book} as $ b_{t}(\mathbf{x})\approx\sum_{i=1}^{N}w^{i}_{t}\delta(\mathbf{x}-\mathbf{x}^{i}_t)$ where $\delta(\mathbf{x})$ denotes the Dirac delta mass located at $ \mathbf{x} $. 

\section{Feedback Policy Calculation}
In this section, we provide the method that we use for calculating the feedback policy.

\subsection{The General Problem}
\textit{Optimization problem:} Given an initial belief state $ b_{t'} $ at time $ t' $ and a goal state $ \mathbf{x}_g $, solve the following optimization problem:
\begin{subequations}\label{eq:main Problem}
\begin{align}
\nonumber \min_{\mathbf{u}_{t':t'+K-1}}&=\sum_{t=t'}^{t'+K-1}\mathbb{E}[c(b_t,\mathbf{u}_t)]
\\ s.t.~b_{t+1}&=\tau(b_{t},\mathbf{u}_t,\mathbf{z}_{t+1})
\\\mathbf{x}_{t+1}&=f(\mathbf{x}_{t},\mathbf{u}_{t},\boldsymbol{\omega}_{t})
\\\mathbf{z}_{t}&=h(\mathbf{x}_{t})+\boldsymbol{\nu}_{t}
\\f_j(\mathbf{x})&>0,~ j\!\in\![1,n_c]
\\g(b&_{t'+K},\mathbf{x}_g)=0
\end{align}
\end{subequations}
where $ c(\cdot,\cdot):\mathbb{B}\times\mathbb{U}\rightarrow\mathbb{R} $ is the one-step cost function, $ \tau:\mathbb{B}\times\mathbb{U}\times\mathbb{Z}\rightarrow\mathbb{B} $ denotes the belief dynamics, $ f_j(\mathbf{x})>0 $ are $ n_c $ number of inequality constraints, and $ g:\mathbb{B}\times \mathbb{X}\rightarrow\mathbb{R} $ denotes the terminal constraint.

\begin{figure}[!t]
   \includegraphics[width=3.5in]{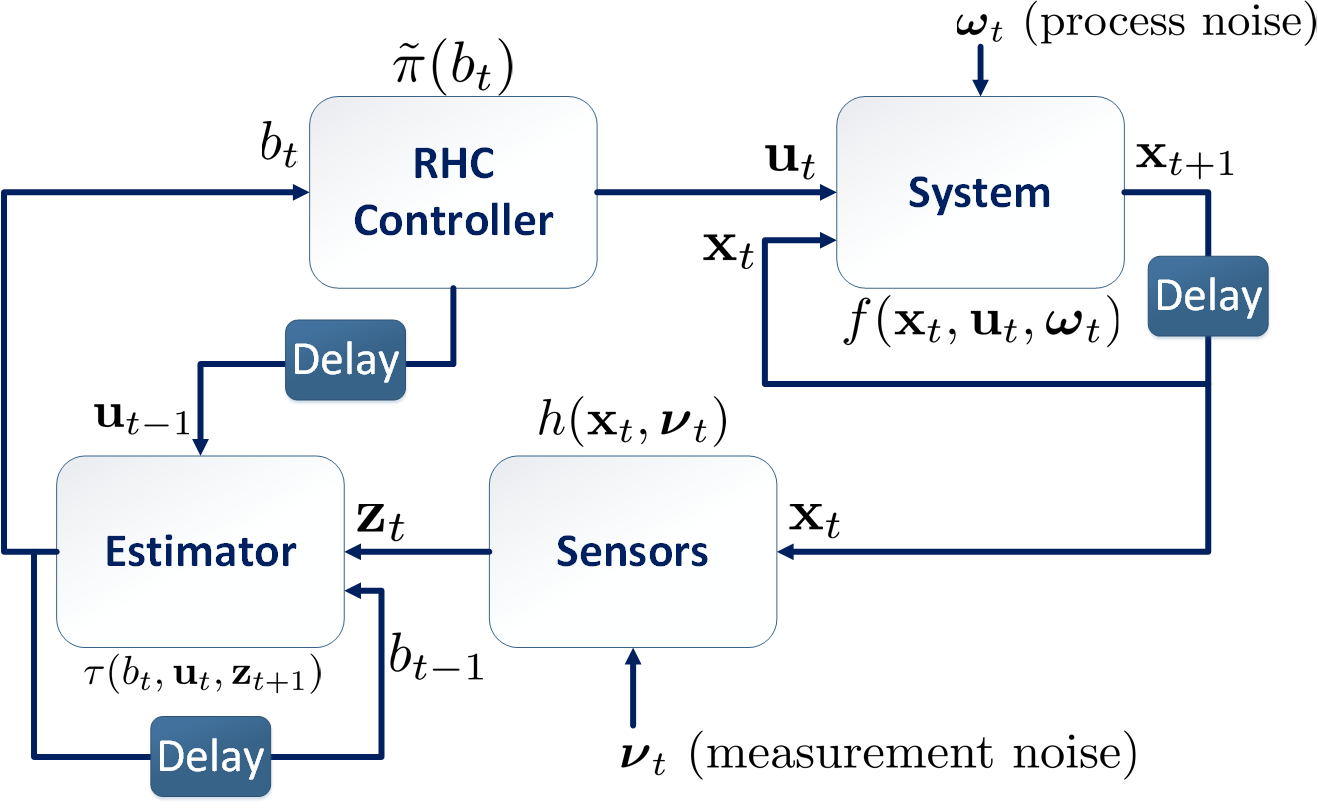}

  \caption{The RHC control loop.}
  \label{fig:RHC_Loop}
\end{figure}

\textit{RHC Strategy:} The control loop is shown in Fig. \ref{fig:RHC_Loop}. The RHC policy function $ \tilde{\pi}:\mathbb{B}\rightarrow\mathbb{U} $ generates an action $ \mathbf{u}_{t}=\tilde{\pi}(b_{t}) $ which is the first element of the sequence of actions generated in problem \eqref{eq:main Problem}. Once $ \mathbf{u}_{t} $ is executed, the state of the system transitions from $ \mathbf{x}_{t} $ into a new state $ \mathbf{x}_{t+1} $ and the sensors perceive a new measurement $ \mathbf{z}_{t+1} $. Given the updated data history, the estimator updates the belief as $ b_{t+1}=\tau(b_{t},\mathbf{u}_t,\mathbf{z}_{t+1}) $. The new belief state is fed into the controller and the loop closes.

\textit{Stopping criteria:} The execution of the actions stop when we reach a belief state $ b $ satisfying:
\begin{align}\label{eq:objective}
\mathcal{P}({b},r,\mathbf{x}_g):=\int_{\mathbf{x}\in\mathit{B}_r(\mathbf{x}_g)}{b}(\mathbf{x})d\mathbf{x}\ge \breve{w}_{th}
\end{align}
where, $\mathit{B}_r(\mathbf{x}_g)$ is defined as $\mathit{B}_r(\mathbf{x}_g):=\{\mathbf{x}\in \mathbb{X}~~\!\! s.t.~~\!\! |\!|\mathbf{x}-\mathbf{x}_g|\!|_2<r, r>0\}$ which is an $r$-ball (in $ L_2 $ norm) centered at $\mathbf{x}_g$, and $0<\breve{w}_{th}<1$ is an arbitrary threshold \cite{Platt-Cvx-RHC-12}.

Now that we have defined our problem, we outline our proposed approach to solve it.

\subsection{Defining the cost function:}
In this subsection, we specifically define the cost function that we use in our optimization problem.

\textit{Maximum-A-Posteriori estimate:} In our optimization problem, we plan for the Maximum-A-Posteriori (MAP) state estimate $\mathbf{x}^{\text{\emph{\tiny MAP}}}_t= \arg\max_{\mathbf{x}\in \mathbb{X}}b_t(\mathbf{x}) $.

\textit{Quadratic cost:} Let us define the one-step cost function $ c(\cdot,\cdot) $ in the optimization problem as follows:
\begin{align}\label{eq:incorporate I}
\!\!\!c(b_t,\mathbf{u}_t)\!\!=\!\!\mathbb{E}_{b_t}[(\mathbf{x}_t-\mathbf{x}^{\text{\emph{\tiny MAP}}}_{t})^T\mathbf{W}_t^{x}(\mathbf{x}_t-\mathbf{x}^{\text{\emph{\tiny MAP}}}_{t})]\!+\!\mathbf{u}_t^T\mathbf{V}_t^{u}\mathbf{u}_t,
\end{align}
where $ \mathbf{W}_t^{x}\succ 0 $, and $ \mathbf{V}_t^{u}\succ 0 $ are weight matrices. Their relative magnitudes incorporate the trade-off between the control effort and the cost of uncertainty in the system. However, unlike the usual Linear Quadratic Regulator (LQR) controller, the weight matrices can be state and time dependent. In this paper, we assume $ \mathbf{V}_t^{u} $ to be constant and design $\mathbf{W}_t^{x} $.

\subsection{Linearizing System Equations}

\textit{Linearized process model:} Given a nominal trajectory of states and controls $ \{\mathbf{x}^{p}_{t}\}_{t=t'}^{t'+K} $, and $ \{\mathbf{u}^{p}_{t}\}_{t=t'}^{t'+K-1} $, we linearize the $ f(\cdot,\cdot, \cdot) $ about the nominal trajectory as follows:
\begin{align}\label{eq:linearized system1}
\nonumber\mathbf{x}_{t+1}&=f(\mathbf{x}^{p}_{t},\mathbf{u}^{p}_t,0)+\mathbf{A}_t(\mathbf{x}_t-\mathbf{x}^{p}_{t}) + \mathbf{B}_t(\mathbf{u}_t-\mathbf{u}^{p}_{t}) +\mathbf{G}_t\boldsymbol{\omega}_t\\
&=f^{p}_{t}+\mathbf{A}_t\mathbf{x}_t + \mathbf{B}_t\mathbf{u}_t +\mathbf{G}_t\boldsymbol{\omega}_t
\end{align}
where $ f^{p}_{t}:= f(\mathbf{x}^{p}_{t},\mathbf{u}^{p}_t,0) -\mathbf{A}_t\mathbf{x}^{p}_{t} - \mathbf{B}_t\mathbf{u}^{p}_{t}$, and $ \mathbf{A}_t=\partial f(\mathbf{x},\mathbf{u},0)/\partial \mathbf{x}|_{ \mathbf{x}^{p}_{t}, \mathbf{u}^{p}_{t} } $, $ \mathbf{B}_t=\partial f(\mathbf{x},\mathbf{u},0)/\partial \mathbf{u}|_{ \mathbf{x}^{p}_{t}, \mathbf{u}^{p}_{t} } $, and $ \mathbf{G}_t=\partial f(\mathbf{x},\mathbf{u}_t,0)/\partial \boldsymbol{\omega}|_{ \mathbf{x}^{p}_{t}, \mathbf{u}^{p}_{t} } $ are the corresponding matrices.

\textit{Linearized observation model:} We are linearizing the $ h(\cdot) $ about the MAP state $ \mathbf{x}^{\text{\emph{\tiny MAP}}}_{t} $ as follows:
\begin{align}
\mathbf{z}_{t}&=h(\mathbf{x}^{\text{\emph{\tiny MAP}}}_{t})+\mathbf{H}(\mathbf{x}^{\text{\emph{\tiny MAP}}}_{t})(\mathbf{x}_t-\mathbf{x}^{\text{\emph{\tiny MAP}}}_{t})+\boldsymbol{\nu}_t,\label{eq:linear observation}
\end{align}
where, $ \mathbf{H}(\mathbf{x}^{\text{\emph{\tiny MAP}}}_{t})=\partial h(\mathbf{x},0)/\partial \mathbf{x}|_{ \mathbf{x}^{\text{\emph{\tiny MAP}}}_{t} } $. Thus, $ \mathbf{H}(\mathbf{x}^{\text{\emph{\tiny MAP}}}_{t}) $ is not a constant matrix, rather it is a function of $ \mathbf{x}^{\text{\emph{\tiny MAP}}}_{t} $ which is a function of control variables in the optimization problem.

\subsection{Incorporating Filtering in the Cost}
Following the Dual Control concept in \cite{Kumar-book-86}, we use the time-and-trajectory-dependent matrix $ \mathbf{W}_t^{x}:=\mathbf{W}(\mathbf{x}^{\text{\emph{\tiny MAP}}}_{t})=\mathbf{H}(\mathbf{x}^{\text{\emph{\tiny MAP}}}_{t})^{T}\mathbf{R}(\mathbf{x}^{\text{\emph{\tiny MAP}}}_{t})\mathbf{H}(\mathbf{x}^{\text{\emph{\tiny MAP}}}_{t}) $ as our weight matrix, where $ \mathbf{R}:\mathbb{X}\rightarrow\mathbb{R}^{n_z\times n_z} $ is a proper weighting matrix, later to be defined. By assigning this weight matrix, we can show that the first term in \eqref{eq:incorporate I}  is equivalent to $ \mathbb{E}[(\mathbf{z}_{t}-\mathbf{z}_{\mathbf{x}^{\text{\emph{\tiny MAP}}}_{t}})^T\mathbf{R}(\mathbf{x}^{\text{\emph{\tiny MAP}}}_{t})(\mathbf{z}_{t}-\mathbf{z}_{\mathbf{x}^{\text{\emph{\tiny MAP}}}_{t}})] $, which is the expected weighted innovation. Choosing this cost, we drop the filtering equation in \eqref{eq:main Problem}, which enables us to achieve the reduced complexity in our solution. Particularly, if  $ \mathbf{R}=\mathbf{I}_{n_z} $ where $ \mathbf{I}_{n_z} $ is the $ n_z $-dimensional identity matrix, the first term in \eqref{eq:incorporate I} is $ \mathrm{trace}(\mathbf{Cov}[(\mathbf{z}_{t}-\mathbf{z}_{\mathbf{x}^{\text{\emph{\tiny MAP}}}_{t}})]) $ where $ (\mathbf{z}_{t}-\mathbf{z}_{\mathbf{x}^{\text{\emph{\tiny MAP}}}_{t}}) $ is the predicted error of the observation at time step $ t $ from its nominal observation. Therefore, conceptually, the minimum of this cost occurs over the state trajectories along which the dispersion in the ensemble of the observation trajectories is reduced in the sense of covariance. This means that the minimization seeks for reducing the uncertainty in the predicted observation or equivalently in estimation. Later, we explain that $ \mathbf{R} $ can be designed suitably to define a more tractable optimization problem.

\subsection{Terminal Constraint}
To reach a specific goal state while minimizing the uncertainty over the trajectory (localization) we set the terminal constraint in equation \eqref{eq:main Problem} as $ \mathbf{x}^{\text{\emph{\tiny MAP}}}_{t'+K} = \mathbf{x}_{g} $.

\section{Particle representation and the cost}
Calculation of the expectation in \eqref{eq:incorporate I} is intractable in general. That is why, we use particle representation to approximate this expectation, and overcome the intractability in solving the problem. Therefore, we can write the overall cost in \eqref{eq:incorporate I} as:
\begin{align}\label{eq:objective with real particles}
\!\!\!\!\!\!\!\sum\limits_{t=t'+1}^{t'+K}\!\![\frac{1}{N}\!\sum_{i=1}^{N}[(\mathbf{x}^i_t-\mathbf{x}^{\text{\emph{\tiny MAP}}}_{t})^T\mathbf{W}(\mathbf{x}^{\text{\emph{\tiny MAP}}}_{t})(\mathbf{x}^i_t-\mathbf{x}^{\text{\emph{\tiny MAP}}}_{t})\!]\!+\!\mathbf{u}_{t-1}^T\mathbf{V}_t^u\mathbf{u}_{t-1}\!],\!\!\!
\end{align}
where $ \{\mathbf{x}^i_t\}_{i=1}^{k} $ are the set of particles obtained through the particle filtering after taking action $ \mathbf{u}_{t-1} $ followed by perceiving $ \mathbf{z}_t $. 
\subsection{Predicting the Evolution of the Particles}\label{subsec:Deterministic Evolution of The Particles Over The Horizon}

We use the linear model of \eqref{eq:linearized system1} to predict the evolution of the particles in \eqref{eq:objective with real particles}. Using the noiseless equations and given initial set of particles $\{\mathbf{x}^{i}_{t'}\}^{k}_{i=1}$ at time step $ t' $, we can iteratively write:
\begin{align*}
\!\!\!\!\!\mathbf{x}^{i}_{t'+t+1}\!\!-\!\!\mathbf{x}^{\text{\emph{\tiny MAP}}}_{t'+t+1}\!\!=\!\!\tilde{\mathbf{A}}_{t':t'+t}(\mathbf{x}^{i}_{t'}-\mathbf{x}^{\text{\emph{\tiny MAP}}}_{t'}),\!\label{eq:lin-sys-evolve-for-convex-derivation-d}
\end{align*}
where $ \tilde{\mathbf{A}}_{t_1:t_2}:=\prod^{t_2}_{\tau=t_1}\mathbf{A}_{\tau} = \mathbf{A}_{t_2}\mathbf{A}_{t_2-1}\cdots\mathbf{A}_{t_1} $, for $ t_1\le t_2 $, otherwise (i.e., for $ t_1>t_2 $), $ \tilde{\mathbf{A}}_{t_1:t_2}:=\mathbf{I}_{n_x} $.

\textit{Simplified cost function:} Let us define a vector $ \mathbf{c}_t := (\mathbf{c}^{1^{T}}_{t}, \mathbf{c}^{2^{T}}_{t}, \cdots, \mathbf{c}^{N^{T}}_{t})^T\in\mathbb{R}^{Nn_x} $, where $ \mathbf{c}^{i}_{t} := \frac{1}{N}\tilde{\mathbf{A}}_{t':t-1}(\mathbf{x}^i_{t'}-\mathbf{x}^{\text{\emph{\tiny MAP}}}_{t'})\in\mathbb{R}^{n_x} $ for $ 1\le i\le N $. Moreover, define a matrix $ \bar{\mathbf{W}}(\mathbf{x}^{\text{\emph{\tiny MAP}}}_{t}) := \mathrm{BlockDiag}(\mathbf{W}(\mathbf{x}^{\text{\emph{\tiny MAP}}}_{t})) $ a block-diagonal matrix with $ N $ equal diagonal blocks of $ {\mathbf{W}}(\mathbf{x}^{\text{\emph{\tiny MAP}}}_{t}) $. Thus, the simplified cost is:
\begin{align}
\sum\limits_{t=t'+1}^{t'+K}[\mathbf{c}^{T}_t\bar{\mathbf{W}}(\mathbf{x}^{\text{\emph{\tiny MAP}}}_{t})\mathbf{c}_t+\mathbf{u}_{t-1}^T\mathbf{V}_t^u\mathbf{u}_{t-1}],
\end{align}
where $ \mathbf{c}_t $ is a constant vector at time step $ t $ and is defined as before. Moreover $ \mathbf{x}^{\text{\emph{\tiny MAP}}}_{t} = \tilde{\mathbf{A}}_{t':t-1}\mathbf{x}^{\text{\emph{\tiny MAP}}}_{t'}+\sum_{s=t'}^{t-1}\tilde{\mathbf{A}}_{s+1:t-1}(\mathbf{B}_s\mathbf{u}_{s}+f_s^p) $.

\subsection{Convexity of the Cost}\label{subsec:Convexity of the Cost}
In this subsection we go through the necessary steps to design the convex problem whose solution is the optimized trajectory over the convex feasible space.

\lemma\label{lemma 1} Given the smooth differentiable function $ h(\mathbf{x}):\mathbb{X}\rightarrow\mathbb{R} $, define $ l:\mathbb{X}\rightarrow\mathbb{R} $, as
\begin{align*}
l(\mathbf{x}):=\sqrt{R(\mathbf{x})}\sum\limits_{i=1}^{n_x}d_i\frac{\partial h(\mathbf{x})}{\partial x_i},
\end{align*}
where $\mathbf{x}=(x_1,\cdots,x_{n_x})^T\in\mathbb{X}$, and $ \mathbf{d}=(d_1,\cdots,d_{n_x})^T\in\mathbb{R} $ is an arbitrary vector. If $ l $ is convex or concave in $ \mathbf{x} $, then $ g:\mathbb{X}\rightarrow\mathbb{R}_{\ge 0} $, where $ g(\mathbf{x}):=\mathbf{d}^T\mathbf{H(x)}^TR(\mathbf{x})\mathbf{H(x)}\mathbf{d} $ is a convex function of $ \mathbf{x} $, where $ \mathbf{H}(\mathbf{x}) $ is the Jacobian of $ h $.

\noindent The hint to prove this lemma is to show that $ g(\mathbf{x}) = (l(\mathbf{x}))^2 $.

Let us provide two most common observation models in the literature and design $ R $ for them.

\example \label{example 1} \textit{Range-based measurements:} Let $ h(\mathbf{x}) = |\!|\mathbf{x}-\mathbf{L}|\!|_2$, where $ |\!|\cdot|\!|_2 $ denotes the Euclidean norm, and $ \mathbf{L}\in \mathbb{R} $ represents a landmark. Using $ {R}(\mathbf{x}) = |\!|\mathbf{x}-\mathbf{L}|\!|_2^2$, $ g(\mathbf{x}) $ of Lemma \ref{lemma 1} becomes convex in $ \mathbf{x} $.

\example \textit{Bearing-based measurements:} Given state vector $ \mathbf{x}=[x, y, \theta]^T $, and $ \mathbf{L}=[L_x,L_y]^T $, using $ {R}(\mathbf{x})= (x-L_x)^2+(y-L_y)^2 $ , $ g(\mathbf{x}) $ of Lemma \ref{lemma 1} becomes convex in $ \mathbf{x} $.

The results of Lemma \ref{lemma 1} can be easily extended to the cases where there are multiple observations. For our purposes, we design $ \mathbf{R} $ to be a positive and increasing function of the distance from the landmarks, such that it helps to maintain the convexity of the problem as desired in Lemma \ref{lemma 1}. Therefore, as the state gets distant from the landmarks, the distance gets more penalized, and the corresponding observation bundle is more penalized, as well.

\section{Non-convex Constraints on the State}\label{sec:Non-convex Constraints on the State}

In this section, we provide our solution for handling non-convex constraints on the state. Particularly, we consider the presence of non-convex obstacles which make the feasible state space non-convex. We define a special type of penalty functions to softly incorporate the non-convex constraints in the optimization cost function.

\textit{Constraints approximated by ellipsoids:} Let us consider $ n_c $ number of non-convex constraint $ f_i(\mathbf{x})>0 $, for $ i=1,\cdots,n_c $. Given $ \mathbf{x}=(x_1, \cdots,x_{n_x})^{T}\in\mathbb{X} $, the compact volume constrained by $ f_i(\mathbf{x})\le 0 $ can be covered by a finite number, $ n_b $, of ellipsoids with different areas:
\begin{align*}
f'_i(\mathbf{x}):=\sum_{j=1}^{n_x}-\alpha_{ij}(x_j-c'_{ij})^2,~i=1,\cdots,n_b
\end{align*}
where $ \mathbf{c}'_i:=(c'_{i1}, c'_{i2},\cdots, c'_{in_x})^{T}\in\mathbb{R}^{n_x} $, $ \boldsymbol{\alpha}_i:=(\alpha_{i1}, \alpha_{i2},\cdots, \alpha_{in_x})^{T}\in\mathbb{R}^{n_x} $ are parameters of these ellipsoids. New set of constraints \{$ f'_i(\mathbf{x})>0, ~i=1,\cdots,n_b $\} is at least as conservative as \{$ f_i(\mathbf{x})>0, ~i=1,\cdots,n_c $\}. Moreover, matrices $ \mathcal{C} := (\mathbf{c}'_1, \mathbf{c}'_2,\cdots, \mathbf{c}'_{n_b})\in\mathbb{R}^{n_x\times n_b} $ and $ \mathcal{A} := (\boldsymbol{\alpha}_1, \boldsymbol{\alpha}_2,\cdots, \boldsymbol{\alpha}_{n_b})\in\mathbb{R}^{n_x\times n_b} $ contain all the parameters needed to capture the environment's obstacles.

\textit{Obstacle Penalty Function:} We define an Obstacle Penalty Function (OPF) as follows:
\begin{align}
f^{b}(\mathbf{x}):=\max_{1\le i\le n_b}\{Me^{\sum_{j=1}^{n_x}-\alpha_{ij}(x_j-c'_{ij})^2}\}
\end{align}
where $ M>0 $ is a big penalizing number. By adding this OPF to the optimization cost, the optimization will seek to minimize this cost as well. The OPF can be designed to be nearly zero (of the order of $ 10^{-40} $ or so), except in the vicinity of the area enclosed by the ellipsoid where it sharply gains values. Particularly, $ \boldsymbol{\alpha}_i $, and $ M $ can be chosen appropriately such that the function has well small values in a desired margin of safety outside the banned areas. The nice property of this penalty function is that it is continuous and differentiable infinitely many times, everywhere, which is useful in gradient decent methods. Therefore, if we purely use the gradient decent methods, initialized by trajectory that is not in the local minimal of the OPF, this function will act as a barrier and prevent the trajectory from getting inside the infeasible states.

\textit{RHC inner loop optimization problem:} Given the initial belief state $ b_{t'}(\mathbf{x})=(1/N)\sum_{i=1}^{N}\delta(\mathbf{x}-\mathbf{x}^{i}_t) $ at time $ t' $, a goal state $ \mathbf{x}_g $ and obstacle parameters $ (\mathcal{A},\mathcal{C}) $, solve the following optimization problem:
\begin{align}\label{eq:main Problem_1}
\nonumber \min_{\mathbf{u}_{t':t'+K-1}}&\sum\limits_{t=t'+1}^{t'+K}[\mathbf{c}^{T}_t\bar{\mathbf{W}}(\mathbf{x}^{\text{\emph{\tiny MAP}}}_{t})\mathbf{c}_t+\mathbf{u}_{t-1}^T\mathbf{V}_t^u\mathbf{u}_{t-1}+\beta f^{b}(\mathbf{x}^{\text{\emph{\tiny MAP}}}_{t})]
\\ s.t.~~&\mathbf{x}^{\text{\emph{\tiny MAP}}}_{t'+K}=\mathbf{x}_g
\end{align}
where $ \beta $ is set to zero for solving the convex problem and set to one for solving the problem with obstacles. Moreover, all the parameters and functions are as defined before.

The overall algorithm for the planning problem is expressed in \ref{alg:re-planning}.
\begin{algorithm}[t]
    \SetKwInOut{Input}{Input}

    \Input{Initial belief state $b_{t'}$, Goal state $\mathbf{x}_{g}$, Planning horizon $K$, Belief dynamics $\tau$, Obstacle parameters $ (\mathcal{A},\mathcal{C}) $}
    \While{$\mathcal{P}(b_t,r,\mathbf{x}_g)\le \breve{w}_{th}$}{
    Solve problem \eqref{eq:main Problem_1}\;
    $ \mathbf{u}_{t}\gets\tilde{\pi}(b_{t}) $\;
    execute $ \mathbf{u}_{t} $, perceive $ \mathbf{z}_{t} $\;
    $b_{t+1}(\mathbf{x})\gets\tau(b_{t}(\mathbf{x}),\mathbf{u}_{t},\mathbf{z}_{t})$\;
    }
\caption{Planning Algorithm}\label{alg:re-planning}
\end{algorithm}

\section{Simulations and Examples}\label{sec:simulations}
In this section, we show some applications for our method. We perform all our simulations in MATLAB 2015b in a 2.90 GHz CORE i7 machine with dual core technology and 8 GB of RAM. First, we do a comparison test with an example in the literature and analyze the solutions of two algorithms with various parameters. Then, we introduce a scenario that consists of guiding a robot between two walls. Our last experiment is a simulation where a robot is in a complex scenario in a house with several features to localize with respect to and reach a goal.

\subsection{Comparison Test in a Convex Scenario}
\begin{figure}[ht!]
  \centering
  {\includegraphics[width=2.7in]{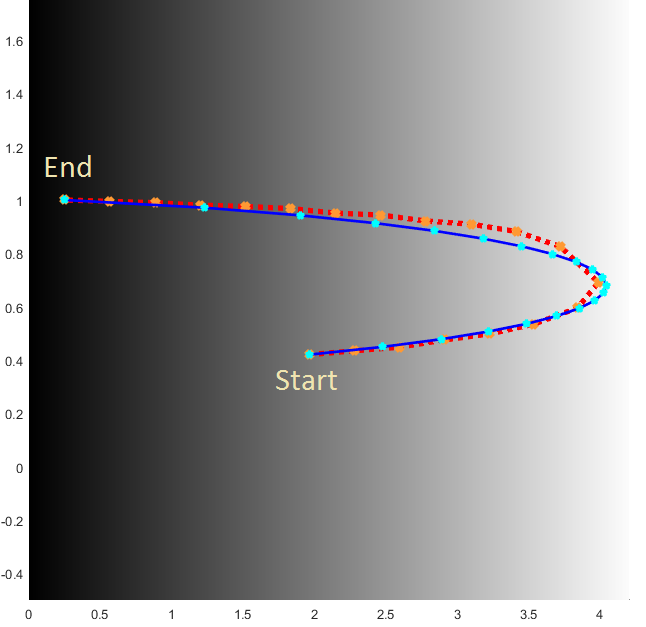}}
  \caption{Light dark example. Lighter states on the right mean presence of less observation noise. The solid blue and red dotted lines show the results of our method and the implementation of \cite{Platt-Cvx-RHC-12}, respectively.\label{fig:Light_dark_our_method_1} }
\end{figure}

In this experiment, we consider the light-dark example introduced in \cite{Platt-Cvx-RHC-12}. We compare our results with the algorithm presented in \cite{Platt-Cvx-RHC-12}. Since we did not have access to the author's code, we implemented the method of \cite{Platt-Cvx-RHC-12} in MATLAB to the best of our ability. Note that in this scenario, we assume that there is no obstacle in the environment. It is important to note that essentially, the two methods are different from each other, however, we solve the same problem for the same systems and same initial and final states. However, the reader cannot find a mapping between the methods. Therefore, the optimization tuning parameters are different and have different meanings. The state, observation and action spaces are 2-dimensional continuous spaces. The process model is linear with $ \mathbf{A}=\mathbf{B}=\mathbf{I}_{2} $, and the observation model is linear with non-linear observation covariance, modeled as $ \mathbf{R}(x)=\mathrm{diag}(1/(2x_1+1),1/(2x_1+1)) $, where $ x_1>0 $ is the first element of state. Therefore, as the robot gets further to bigger values of $ x_1 $ it can localize better with less noisy observations. This is shown in Fig. \ref{fig:Light_dark_our_method_1} with lighter background on th right side. One can verify that the problem is convex in both methods (with different shapes of cost functions). Figure \ref{fig:Light_dark_our_method_1} shows the results of the optimized trajectory for time 0 with 1000 particles and a time horizon of 20. Moreover, to avoid the control saturation, we add a constraint to bound the control input's magnitude at each step to 3.16. The initial distribution is a mixture of two Gaussians with equal variances of 0.0625 and means at (1.75, 0) and (2, 0.5). The solid line shows the results for our problem with $ \mathbf{V}_t^u = 0.065 $. It should be noted that, in our simulation, changes $ \mathbf{V}_t^u $ does not impose unexpected behavior in the trajectory. Rather, by increasing the values of $ \mathbf{V}_t^u $, the agent acts more conservatively in terms of the consumed energy effort.

\textit{Sensitivity of solution to number of particles: }We increase the number of particles from 50 to 1000, 10000, and 100000 particles and analyze the optimization size and required time for the optimization. In our method, by increasing the number of particles, the optimization vector size does not increase. Neither are additional constraints introduced by increasing the number of particles. Therefore, as shown in table \ref{table:comparison} the required time for optimization does not increase significantly. However, in \cite{Platt-Cvx-RHC-12}, the optimization vector size is dependent on the number of particles, particularly, it is equal to $ (Kn_u+N) $, while in our method, it is only $ Kn_u $. Moreover, in their method, by addition of one particle, $ K $ new inequality constraints are added to the optimization problem, whereas in our method, there is no such constraint and the number of optimization constraints is independent from the number of samples. The results of table \ref{table:comparison} show that our method is scalable with the number of particles. As stated in the table, for $ N= 10000 $ and $ N=100000 $, we could not perform the optimization for the method in \cite{Platt-Cvx-RHC-12} due to high required memory allocation.

\textit{Sensitivity of solution to time horizon: } Lastly, we perform the optimization for lookahead time horizon $ K=10, 20, 50$ and  $ 100 $ and report the required time in table. Once again, since the number of optimization variables is $ Kn_u $ which is $ 2K $, and there is no added constraint for addition of time horizon, the optimization time does not explode in our method. Whereas, in \cite{Platt-Cvx-RHC-12}, increasing the time horizon, increases the solution time significantly. The results reflected in table \ref{table:comparison} show that our method is scalable with long time horizon as well. However, for $ K= 50 $ and $ K=100 $, we could not perform the optimization for method of \cite{Platt-Cvx-RHC-12} due to high required memory allocation.
\begin{table*}\centering
\ra{1.3}\caption{The results of our comparative simulations for several time horizons and particle numbers in a convex light-dark scenario.} \label{table:comparison}
\begin{tabular}{@{}c c c c c c c c c c @{}}\toprule
& Time horizon ($ K $) & \multicolumn{4}{c}{20} & 10 & 20 & 50 & 100\\
\cline{2-10}
& Number of Particles ($ N $) & 100 & 1000 & 10000 & 100000 & \multicolumn{4}{c}{1000}  \\ \midrule
	& Time (s) &	0.24	&	0.33	&	1.11 & 10.37	& 0.16&	0.33 & 2.30 & 9.22	\\
	& \# of Iterations &	288	&	288	&	288	&  288 & 170 & 288 &  1013 & 2215	 \\
Our	& Function Tolerance &	2e-03	&	2e-03	&	2e-03	& 2e-03 &	2e-03 &	2e-03 & 2e-03 & 2e-03 \\
Method	& Constraint Tolerance & 5.551e-16	&	8.882e-16	&	5.551e-16 &  2.331e-15	& 1.110e-15 &	8.882e-16 & 3.839e-11 & 1.883e-11 \\
	& \# of Opt. Vars.$ ^{\dagger} $ ($ Kn_u$) &	40	&	40	&	40 & 40	& 20&	40 & 100 & 200	\\
	& \# of Opt. Constrs.$ ^{\dagger} $ ($ 1 $) & 1 & 1 & 1 & 1 & 1 & 1 & 1 & 1	\\
\midrule
    & Time (s) &	49.0 &	311.32 & * &	*	& 80.22 &	311.32 &* & *\\
	& \# of Iterations &	40000	&	40000	&		&  & 40000 & 40000 & &	\\
Method	& Function Tolerance &	2e-02	&	2e-02	&	&  & 2e-02 & 2e-02  & &\\
 of	& Constraint Tolerance &	3.509e-04	&	5.853e-04	&	& & 5.671e-04 & 5.853e-04	&  & 	\\
\cite{Platt-Cvx-RHC-12}	& Required Memory (GB) &		&		& 15.0 & 1490.7	&	 & & 37.6& 76.0	\\
	& \# of Opt. Vars.$ ^{\dagger} $ ($ Kn_u\!\!+\!\!N $) &	140	&	1040	&	10,040 & 100,040	& 1020&	1040 & 1100 & 1200	\\
	& \# of Opt. Constrs.$ ^{\dagger} $ ($ N(K\!\!-\!\!1)\!\!+\!\!1 $) &	1901	&	19,001	&	190,001 & 1,900,001	& 9001 &	19,001 & 49,001 & 99,001	\\
\bottomrule
\end{tabular}\vspace{+5pt}\par
\hspace{-301pt}*: Unable to allocate enough memory to solve the problem.\\
$ ^{\dagger} $: `\# of Opt. Vars.' specifies the number of optimization variables, and `\# of Opt. Constrs.' specifies the number of optimization problem's constraints.
\end{table*}

\begin{figure}[ht!]
  \centering
  {\includegraphics[width=3.5in]{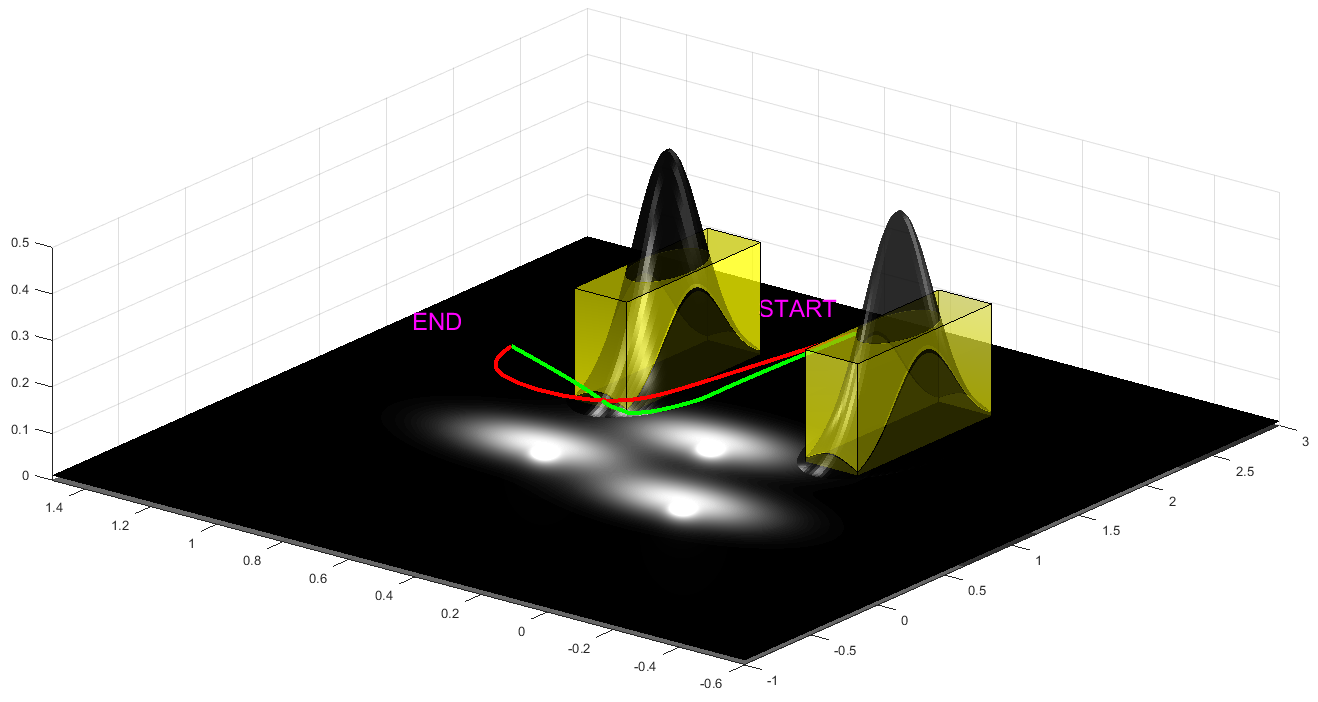}}
  \caption{Robot within two walls. The OPF is visualized within the walls. The green and red lines show the results for optimization with and without considering the walls.\label{fig:two_walls} }
\end{figure}

\subsection{Robot within Two Walls; Visualization of OPF}

In this section, we simulate a case where there is non-convex constraints in the state space. Figure \ref{fig:two_walls} depicts the results in a case where the system starts with a distribution about its initial state and wants to reach the goal state while minimizing the localization error and spending low energy. The green and red lines show the solution of the convex problem where there is no walls, and the problem with added walls, respectively. As it is seen, there are three information sources in that are shown with lighter spots in Fig. \ref{fig:two_walls}. The observation model is range based as described in example \ref{example 1}. To obtain the green trajectory, the convex optimization problem (which is initialized with an arbitrary solution) is solved. Then, the green trajectory (which is not feasible for the case with walls) is used as the initial trajectory for the optimization with OPF to obtain the red trajectory which avoids the walls, as well.

\subsection{Complex scenario}

\begin{figure}[ht!]
  \centering
  {\includegraphics[width=3in]{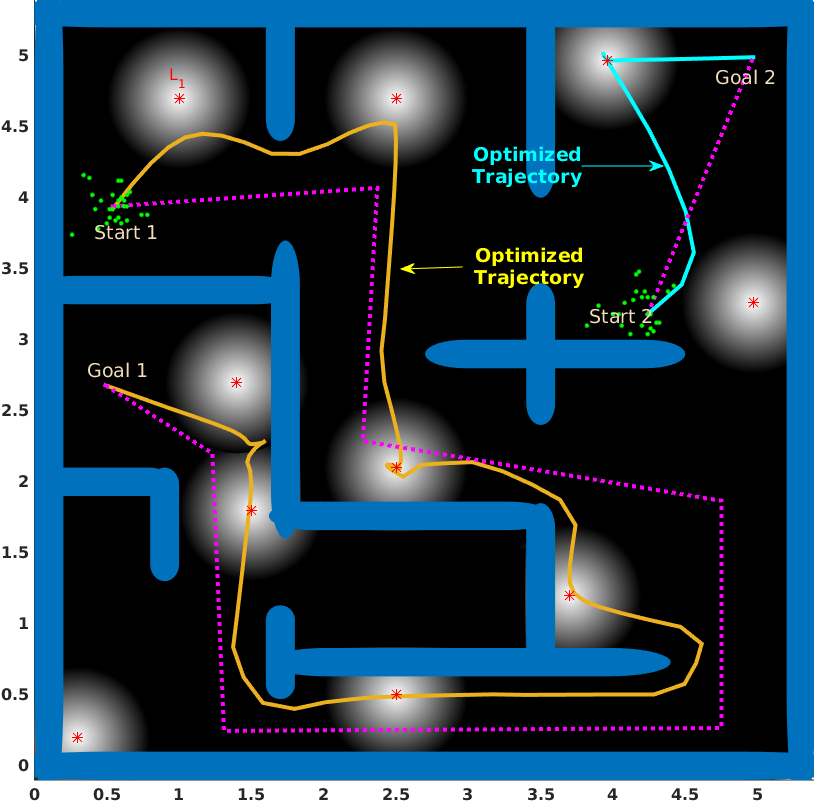}}
  \caption{A holonomic system in a complex scenario. Solid lines show the optimal trajectories, dotted show the initial, for two different scenarios. The longer trajectory includes obstacles, and the other, no obstacles. The dots around the start points show the initial particles. Landmarks are marked as stars and information is coded with color (lighter means more information). Lookahead time horizon for the longer trajectory is 100 and 30 for the other.
\label{fig:paper_12} }
\end{figure}

\textit{Robot in a house:} Figure \ref{fig:paper_12} depicts the results in two cases where the objective is similar to the previous example. In the first case, the robot is put in a room and wants to reach a room in the other side of the house. Given an initial trajectory, shown by red dots, the optimization provides the optimized trajectory that seeks for the information sources in every house, and the penalty functions perform the task of keeping the robot away from the obstacles. In this case, the lookahead time horizon is set to $ K=100 $. In the second case, the start and final goal of the robot is in one room, and therefore, the optimization can solve the problem with any arbitrary trajectory in that room like the straight line.

\section{Conclusion}\label{sec:conclusion}
In this paper, we proposed a method for controlling a stochastic system starting from a configuration in the state space to reach a goal region. Our method consists of a receding horizon control strategy, where planning occurs in the belief space rather than on the underlying state space. Our solution consists of solving a convex program for the unconstrained state space. Hence it can be solved quickly and in an on-line manner where the robot might need to re-plan in case of deviations from the planned trajectory due to the stochastic and noisy nature of the system. Moreover, we proposed a method of incorporating the non-convex constraints in the optimization problem without adding new variables, which enables us to maintain the scalability of our solution for high number of particles and longer lookahead horizons. In our future work, we will extend this method to better process models, and for the cases in which the assumptions can be reliable for a long distance.
\section*{Acknowledgment}
This material is based upon work partially supported by AFOSR Contract No. FA9550-13-1-0008, the U.S. Army Research Office under Contract No. W911NF-15-1-0279, and the
NSF under Contract Nos. CNS-1302182 and Science \& Technology Center Grant CCF-0939370.
\bibliographystyle{IEEEtran}
\bibliography{MohammadRafi}

\end{document}